\documentclass[10pt,letterpaper]{article}
\usepackage[T1]{fontenc}

\usepackage{spconf,amsmath,amssymb,graphicx}
\usepackage{cite,pifont,fontawesome5}
\usepackage{float}
\usepackage{balance}
\usepackage{tikz}
\usetikzlibrary{arrows.meta,positioning,fit,calc}
\usepackage{booktabs,array,etoolbox}
\newcolumntype{P}[1]{>{\raggedright\arraybackslash}p{#1}}
\makeatletter
\AtBeginEnvironment{table}{%
  \apptocmd{\@makecaption}{\vskip 5pt}{}%
    {\PackageError{icassp2027}{Table caption spacing patch failed}{Check spconf.sty}}%
}
\newcommand{\compactfigurecaption}{%
  \patchcmd{\@makecaption}{\vskip 10pt}{\vskip 5pt}{}%
    {\PackageError{icassp2027}{Figure caption spacing patch failed}{Check spconf.sty}}%
}
\AtBeginEnvironment{figure}{\compactfigurecaption}
\AtBeginEnvironment{figure*}{\compactfigurecaption}
\patchcmd{\section}{-3.5ex}{-3.0ex}{}{\PackageError{icassp2027}{Section spacing patch failed}{Check article.cls}}
\patchcmd{\section}{2.3ex}{2.0ex}{}{\PackageError{icassp2027}{Section spacing patch failed}{Check article.cls}}
\patchcmd{\subsection}{-3.25ex}{-2.75ex}{}{\PackageError{icassp2027}{Subsection spacing patch failed}{Check article.cls}}
\makeatother
\makeatletter
\patchcmd{\@maketitle}{\large \bf}{\fontsize{14}{16}\selectfont\bfseries}{}{\PackageError{icassp2027}{Title patch failed}{Check spconf.sty}}
\patchcmd{\@maketitle}{\vskip 2em}{\vskip 2em\vskip 5pt}{}{\PackageError{icassp2027}{Title margin patch failed}{Check spconf.sty}}
\makeatother
\usepackage[protrusion=false]{microtype}
\usepackage{url}
\usepackage{hyperref}
\newcommand{\doi}[1]{\href{https://doi.org/#1}{\nolinkurl{#1}}}
\hypersetup{hidelinks,
  pdftitle={M3D-Net: Hierarchical Coordination of Spatial Context, Feature Reuse, and Differential Attention for Mammography Classification},
  pdfauthor={Zheng Yu; Xinhang Li; Jiabao Gao; Boyang Wang; Xiang Li},
  pdfsubject={Breast image classification with a hierarchical encoder and an adapted image-clinical ultrasound evaluation},
  pdfkeywords={medical image classification, mammography, differential attention}}

\definecolor{paperblue}{HTML}{3A73AA}
\definecolor{paperpurple}{HTML}{8266C7}
\newcommand{\findingbox}[2]{%
  \par\smallskip\noindent
  \begin{tikzpicture}
  \node[draw=#1!65,fill=#1!5,rounded corners=1mm,line width=.5pt,
    inner xsep=4pt,inner ysep=4pt,text width=\dimexpr\columnwidth-9pt\relax,
    align=justify,font=\normalfont\ninept] {#2};
  \end{tikzpicture}\par\smallskip
}
\newcommand{\yes}{\ding{51}}
\newcommand{\no}{\ding{55}}

\title{M3D-Net: Hierarchical Coordination of Spatial Context, Feature Reuse, and Differential Attention for Mammography Classification}

\name{Zheng Yu$^{1}$ \qquad Xinhang Li$^{2}$ \qquad Jiabao Gao$^{1,2}$ \quad Boyang Wang$^{2}$\qquad Xiang Li$^{3,\ast}$}
\address{
$^{1}$Shenzhen Loop Area Institute, Shenzhen, China\\
$^{2}$The Chinese University of Hong Kong, Shenzhen, China\\
$^{3}$Shenzhen Research Institute of Big Data, Shenzhen, China\\
$^{\ast}$Corresponding author: Xiang Li}

\begin{document}

\ninept
\setlength{\abovedisplayskip}{6pt plus 1pt minus 1pt}
\setlength{\belowdisplayskip}{6pt plus 1pt minus 1pt}
\setlength{\abovedisplayshortskip}{3pt plus 1pt}
\setlength{\belowdisplayshortskip}{5pt plus 1pt minus 1pt}
\flushbottom

\maketitle

\begin{abstract}
Breast image classification requires local detail and global tissue context, yet these cues can weaken as representations deepen. We present M3D-Net, a mammography encoder that hierarchically coordinates multi-scale coordinate attention, bounded dynamic feature reuse, and differential attention through resolution-aware operator placement. Within-stage retrieval preserves access to earlier features, coordinate-aware aggregation integrates local and global context, and differential attention operates at coarse resolutions. We evaluate image-only classification on AISSLab mammography and an adapted image--clinical model on BrEaST ultrasound. Against EdgeNeXt, RepViT, and TransXNet, the proposed implementations achieve the highest recorded validation accuracy and late-training accuracy, with the lowest endpoint cross-entropy loss. Validation accuracies reach 97.78\% and 80.39\%, respectively. These results support further evaluation of hierarchical coordination across breast imaging settings; repeated-seed, component-controlled, and independent evaluations remain necessary.
\end{abstract}

\noindent Code is available at \href{https://github.com/YuZhengYYDS/M3D-Net}{\faGithub\enspace\texttt{github.com/YuZhengYYDS/M3D-Net}}\par


\begin{figure*}[!t]
\centering
\includegraphics[width=\textwidth]{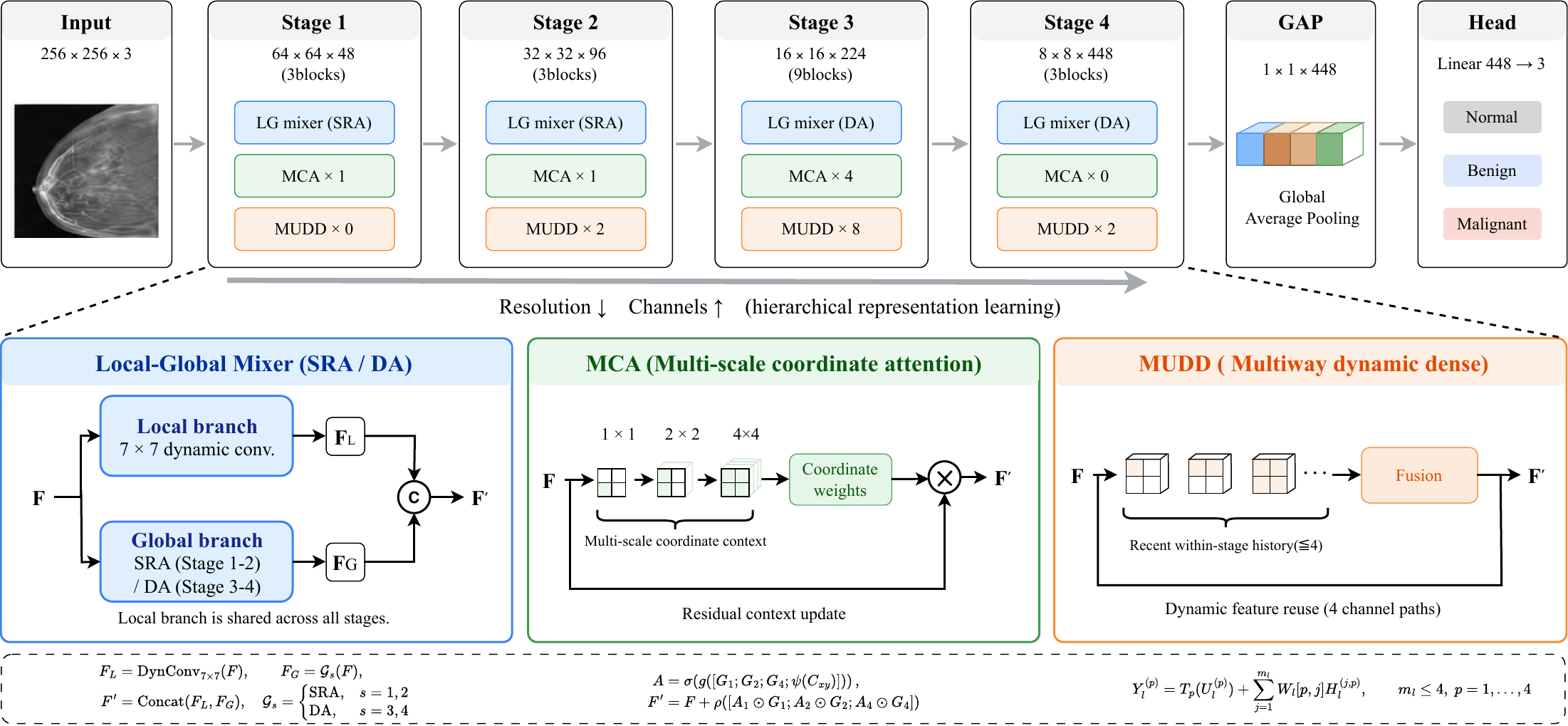}
\caption{Stage-centric overview of the base mammography encoder.
The upper row shows the four-stage hierarchy, GAP, and three-class head. Each stage card gives height$\times$width$\times$channels, block depth, and operator allocation: the LG mixer uses a shared local branch with SRA in Stages~1--2 and DA in Stages~3--4, while MCA and MUDD are activated in the indicated numbers of blocks.
The dashed guides connect these stage-level settings to the lower mechanism summaries for local--global mixing, MCA, and MUDD.
Figure~\ref{fig:block} expands one encoder block; Section~\ref{sec:allocation} specifies the exact allocation.}
\label{fig:architecture}
\end{figure*}

\section{Introduction}
\textbf{Mammography requires local detail and tissue context.} Downsampling reduces direct access to earlier patterns that must be interpreted within broader tissue structure~\cite{geras2017}. Deep learning has established the value of mammographic features for screening~\cite{mckinney2020,lotter2021,wu2020,shen2019}, including evaluation in routine implementation~\cite{eisemann2025}. This motivates coordinating context aggregation, feature reuse, and global mixing.

\textbf{Prior backbones emphasize different design axes.} TransXNet combines dynamic local convolution with global attention~\cite{lou2025}; EdgeNeXt combines convolution with channel-space attention~\cite{maaz2022}; RepViT is a lightweight CNN informed by Transformer design~\cite{wang2024}. Recent mammography work also investigates region-guided token selection and contrastive learning with pretrained vision models~\cite{sanghvi2026}. Table~\ref{tab:positioning} positions M3D-Net against these backbones and Transformer mechanism precedents. Our design therefore aims to preserve fine-grained local cues while integrating broader tissue context as spatial resolution decreases. We pursue this goal through resolution-aware hierarchical coordination of spatial aggregation, within-stage feature reuse, and differential attention.

\begin{table}[!t]
\caption{Comparison of published block designs. Conv.: spatial convolution; Reuse: input-dependent layer history; Diff.: two-map attention subtraction. \yes/\no: present/absent in the cited design.}
\label{tab:positioning}
\centering\small
\setlength{\tabcolsep}{3pt}
\renewcommand{\arraystretch}{1.03}
\begin{tabular}{@{}lcccP{0.33\columnwidth}@{}}
\toprule
\textbf{Method} & \textbf{Conv.} & \textbf{Reuse} & \textbf{Diff.} & \textbf{Primary design}\\
\midrule
EdgeNeXt~\cite{maaz2022} & \yes & \no & \no & Conv. + attention\\
RepViT~\cite{wang2024} & \yes & \no & \no & ViT-inspired CNN\\
TransXNet~\cite{lou2025} & \yes & \no & \no & Dynamic mixing\\
MUDDFormer~\cite{xiao2025} & \no & \yes & \no & Dynamic depth reuse\\
Diff. Tr.~\cite{ye2025} & \no & \no & \yes & Map subtraction\\
\textbf{M3D-Net} & \yes & \yes & \yes & \textbf{Joint allocation}\\
\bottomrule
\end{tabular}
\end{table}

\textbf{M3D-Net assigns complementary roles to three operators.} Multi-scale coordinate attention (MCA) weights local--global outputs; multiway dynamic dense (MUDD) reuse retrieves within-stage history before mixing; differential attention (DA) subtracts global maps at coarse resolution. M3D denotes \emph{multi-scale, multiway, and differential} processing of two-dimensional mammograms.

\textbf{This hierarchical design preserves access to earlier features while integrating local detail with broader tissue context.}
Building on coordinate attention~\cite{hou2021}, dense connections~\cite{huang2017}, MUDDFormer~\cite{xiao2025}, and Differential Transformer~\cite{ye2025}, M3D-Net integrates explicit 2-D coordinates, bounded within-stage channel-path history, and a logit-scaled second attention map through resolution-aware operator placement.

Our hierarchical coordination scheme specifies both \textbf{stage-level allocation} and \textbf{block-level composition}. Across stages, differential attention is restricted to coarse resolutions, while MCA and MUDD follow stage-specific activation schedules. Where enabled, MUDD retrieves recent within-stage features before local--global mixing, whereas MCA weights the mixed features afterward.

\textbf{Our contributions are threefold.} (i) a resolution-aware hierarchical coordination scheme that specifies both stage-level operator allocation and block-level interaction among spatial context weighting, bounded feature reuse, and differential aggregation. (ii) An operator analysis characterizes rejection of constant value components alongside residual feature routes. (iii) We compare mammography backbones and evaluate an adapted image--clinical ultrasound model.

\pagebreak[4]
\section{M3D-Net Architecture}
\textbf{A stage groups blocks at one resolution and channel width.} A stride-four $7\times7$ stem maps a $3\times256\times256$ image to $64\times64$ features. Stages 1--4 have widths $(48,96,224,448)$ and depths $(3,3,9,3)$; stride-two transitions produce resolutions $(64,32,16,8)^2$. This hierarchy retains fine detail early and makes dense global attention affordable later. The scaffold in Fig.~\ref{fig:architecture} follows TransXNet~\cite{lou2025} and ends with global average pooling (GAP) and a linear head for \mbox{three mammography classes}.

\findingbox{paperblue}{\textbf{Resolution-aware design.} Coarse grids reduce attention cost; bounded history retains earlier features from \mbox{the current stage}.}

\begin{figure}[!t]
\centering
\includegraphics[width=\linewidth]{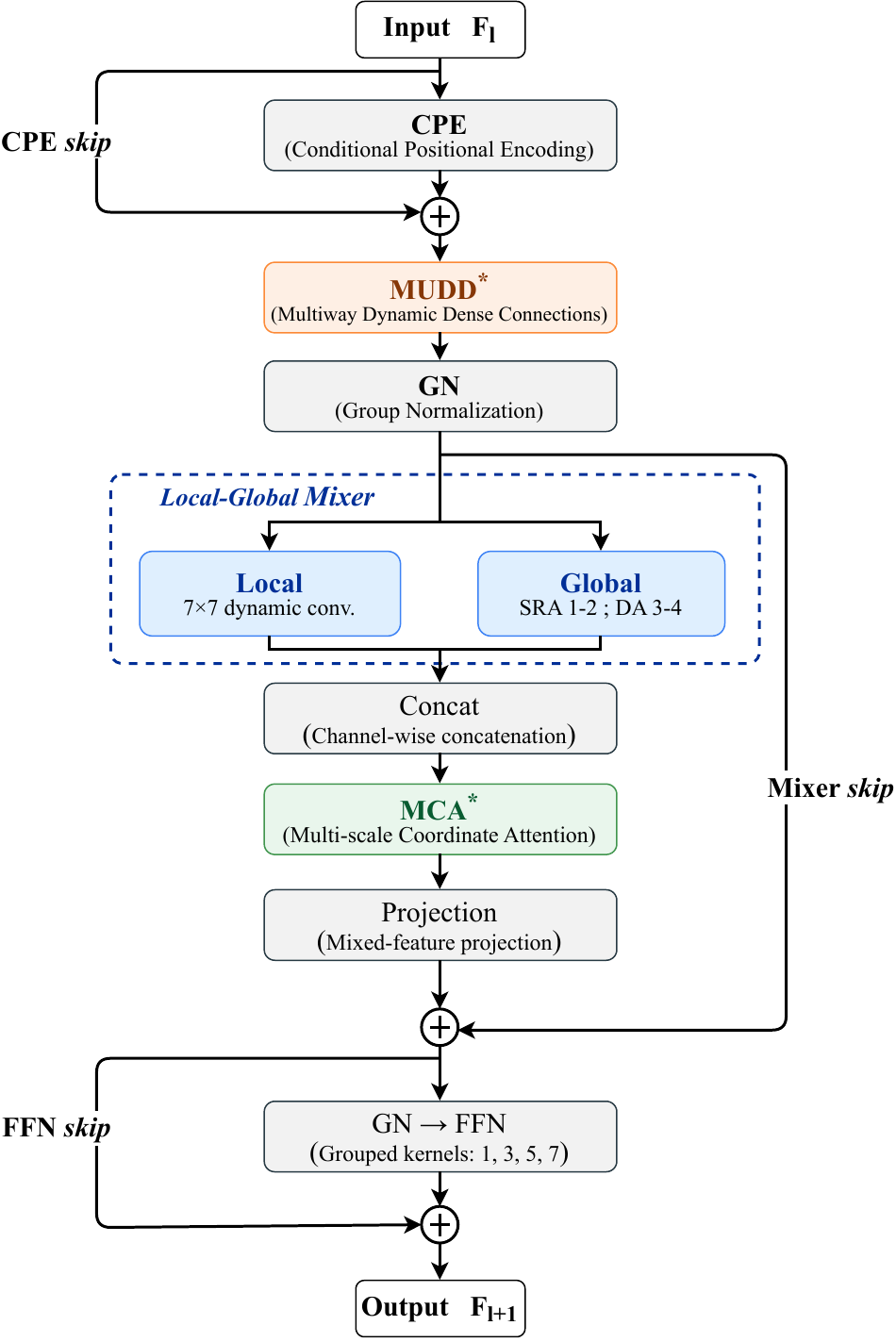}
\caption{Core encoder block of M3D-Net. The block combines CPE, optional MUDD, local--global mixing with dynamic convolution and SRA/DA, optional MCA, and a GN--FFN module. Residual shortcuts bypass CPE, the mixer, and FFN; asterisks indicate stage-dependent activation.
}
\label{fig:block}
\end{figure}

Figure~\ref{fig:block} explains the core unit. Conditional positional encoding (CPE) adds a depthwise $7\times7$ convolution to the input. MUDD retrieves earlier features where active; group normalization (GN) precedes the local--global mixer. The local half uses dynamic $7\times7$ convolution, and the global half uses spatially reduced attention (SRA) or DA. Concatenation, optional MCA and projection complete the mixer. A GN--feed-forward network (FFN) follows. Three residual routes preserve the input to CPE, the mixer and the FFN~\cite{he2016}; the latter two use learned channelwise scaling. The FFN mixes grouped kernels of sizes $1,3,5,7$ after \mbox{fourfold channel expansion}. Stage and block indices fix operator activation; learned gates and history weights depend on \mbox{the current input}.\par

\subsection{Context: multi-scale coordinate attention for local-global integration}
\textbf{MCA adds coordinate-aware pooled context while retaining its input.} Following coordinate and pyramid-pooling precedents~\cite{hou2021,zhao2017}, it divides $F$ into three channel groups for grids $s\in\{1,2,4\}$:
\begin{equation}
G_s=\operatorname{Up}_{H,W}\!\left(\phi_s(\operatorname{Pool}_{s\times s}(F^{(s)}))\right).
\end{equation}
Here $\phi_s$ is pointwise convolution, batch normalization and GELU; bilinear upsampling restores the input grid. Normalized 2-D coordinates $C_{xy}$ condition the update:
\begin{align}
A &= \sigma\!\left(g([G_1;G_2;G_4;\psi(C_{xy})])\right),\\
\operatorname{MCA}(F) &= F+\rho([A_1\odot G_1;A_2\odot G_2;A_4\odot G_4]).
\end{align}
Pointwise network $g$ predicts bottleneck weights, $\psi$ embeds coordinates, and $\rho$ projects and normalizes \mbox{the weighted groups}.

\subsection{Multiway dynamic dense connections}
\textbf{MUDD retrieves at most four earlier within-stage features.} Dense connections~\cite{huang2017} and MUDDFormer~\cite{xiao2025} motivate direct, input-dependent cross-layer access. A pooled bottleneck predicts a row-softmax $4\times4$ matrix $W_l$. Four channel paths update the CPE output $U_l$ as
\begin{equation}
Y_l^{(p)}=T_p(U_l^{(p)})+\sum_{j=1}^{m_l}W_l[p,j]H_l^{(j,p)},\quad m_l\leq4.
\end{equation}
Each $T_p$ is depthwise $3\times3$ convolution, GELU and pointwise projection. The concatenated paths enter the mixer. History stores recent outputs chronologically and resets at stage transitions. Unused weight columns are dropped without renormalization; retrieval is skipped when history is empty.

\subsection{Differential global attention}
\begingroup
\predisplaypenalty=0
DA splits global queries, keys and values into two groups; $d$ is the head width before splitting:
\begin{align}
A_1 &= \operatorname{softmax}(Q_1K_1^\top/\sqrt d),\\
A_2 &= \operatorname{softmax}(\lambda_h\,\operatorname{GN}(Q_2)K_2^\top/\sqrt d),\\
D &= A_1-A_2,\qquad O_i=DV_i\quad(i=1,2).
\end{align}
\endgroup
The learned per-head $\lambda_h$ starts at $0.5$; GN acts on stacked second-query groups. Outputs are concatenated and projected. This adapts two-map differential attention~\cite{ye2025}. \textbf{With attention dropout disabled, $D\mathbf1=0$.} Thus $DV=0$ for spatially constant values $V=\mathbf1c^\top$ before output projection. This operator property does not imply full-encoder invariance or artifact suppression.

\subsection{Resolution-aware operator allocation}
\label{sec:allocation}
Stages 1--4 use $(1,2,4,8)$ global heads. MCA acts in the first $(1,1,4,0)$ blocks; MUDD performs $(0,2,8,2)$ nonempty-history updates. SRA handles stages 1--2, while DA handles all nine and three blocks of stages 3--4. Each DA head stores $2N^2$ map entries at only $N=256$ or $64$ tokens; bounded history holds at most four feature maps. This explains the stage hierarchy's computational role; throughput and parameter-matched efficiency remain unmeasured.

\section{Experiments, Results, and Analysis}
\subsection{Datasets and inputs}
\textbf{We evaluate two complementary settings: image-only mammography classification and image--clinical ultrasound classification.} AISSLab contains 266 mammograms: 100 normal, 66 benign and 100 malignant images. Images came from its Kaggle release~\cite{kaggle_aisslab}; the source describes radiologist-derived annotations~\cite{aisslab2026}. We evaluate three-class classification of these labels, not pathology-confirmed diagnosis or prospective screening. This experiment used the image encoder described in Section~2.

\begin{table}[!t]
\caption{Mammography validation. Acc.: epoch-50 accuracy; Tail: mean accuracy over epochs 41--50; loss: epoch-50 validation loss. Bold marks the best value in each column.}
\label{tab:comparison}
\centering\small
\setlength{\tabcolsep}{4pt}
\begin{tabular}{lrrr}
\toprule
Model & Acc. (\%) & Tail (\%) & Val. loss\\
\midrule
EdgeNeXt & 91.30 & 90.76 & 0.1727\\
RepViT & 93.33 & 92.80 & 0.1514\\
TransXNet & 95.19 & 94.70 & 0.1300\\
\textbf{M3D-Net} & \textbf{97.78} & \textbf{97.22} & \textbf{0.0542}\\
\bottomrule
\end{tabular}
\end{table}

BrEaST~\cite{pawlowska2024breast} contains one ultrasound image per patient for 256 patients. The binary target groups 98 malignant cases against 158 nonmalignant cases (154 benign and four normal). Its source reference standards include biopsy or follow-up. The split was fixed before training: 153 training, 51 validation and 52 reserved test patients, with no patient overlap. \textbf{All four models shared this split.} Complete images were used without ground-truth-mask cropping.

BrEaST inputs additionally included 22 clinical encodings covering age, history, symptoms, examination findings, tissue composition and missingness indicators. Age statistics came only from training patients with known age; missing ages used the training mean. Breast Imaging Reporting and Data System (BI-RADS) risk assessments were excluded from both experiments' inputs; BrEaST predictors also excluded pathology diagnosis and target labels.

\subsection{BrEaST adaptation and training}
\label{sec:breast_protocol}
\textbf{The ultrasound comparison uses a shared image--clinical fusion framework.} An adapted iTransformer clinical encoder~\cite{liu2024itransformer} produced 1000 features, concatenated with 1000 image-backbone features. The common head used dimensions $2000\rightarrow1024\rightarrow512\rightarrow2$, GELU and dropout $0.3$. EdgeNeXt, RepViT and TransXNet denote their XXS, M0.9 and T variants, respectively.

\textbf{M3D-Net$^\dagger$ denotes the adapted ultrasound implementation.} It corrects the DA output tensor's channel/spatial arrangement and adds the STE projection shortcut $P(x)=\operatorname{Proj}(x)+x$ inside the token mixer. Both changes alter forward computation without adding learned parameters. Consequently, this experiment is neither a replication of the unmodified mammography implementation nor a test of a frozen mammography-trained checkpoint.

\textbf{All models trained for 50 epochs with batch size 4 and seed 42.} AdamW used learning rate $2\times10^{-5}$, weight decay $10^{-4}$, betas $(0.9,0.999)$ and a cosine schedule. Training used unweighted cross entropy, gradient-norm clipping at 3.0 and BF16 mixed precision; validation used FP32. Aspect-preserving resizing and padding produced $256\times256$ inputs. Training added padding-4 random crops, horizontal/vertical flips, $\pm3^\circ$ rotations and color jitter. Validation used no random augmentation and applied ImageNet normalization.

Sample repetition with $R=2$ yielded 76 updates per epoch and 3800 in total; repeated records were not additional patients. Compatible official ImageNet weights initialized the image branches. M3D loaded only matching names and shapes; new modules, clinical encoders and fusion heads were randomly initialized. Pretraining coverage differed across models, but \mbox{all parameters were optimized}.

\subsection{Mammography comparison and trajectories}
\findingbox{paperblue}{\textbf{Complementary evidence.} Endpoint accuracy, late-training performance and validation loss characterize different aspects of model behavior. Their joint interpretation provides a broader comparison than a single checkpoint score.}

\begin{figure}[!t]
\centering
\includegraphics[width=0.99\columnwidth,trim=0bp 7.5bp 0bp 7.5bp,clip]{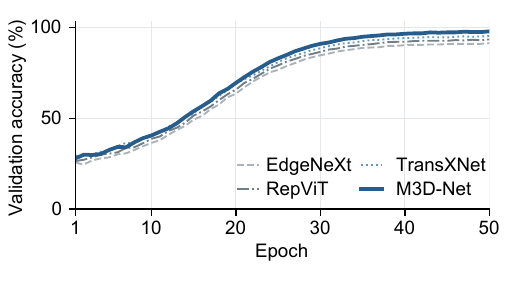}
\caption{Mammography validation trajectories over 50 epochs. Curves retain the recorded values; Tail summarizes epochs 41--50.}
\label{fig:trajectory}
\end{figure}

\textbf{The advantage persists over the final ten epochs.} Figure~\ref{fig:trajectory} shows unsmoothed validation trajectories. M3D-Net averages 97.22\% over epochs 41--50 versus TransXNet's 94.70\%. It first reaches 90\% at epoch 30, compared with epochs 32, 35 and 40 for TransXNet, RepViT and EdgeNeXt. This compares progress in epochs; wall-clock efficiency was not measured.

\subsection{BrEaST validation results}
\label{sec:breast_results}
\findingbox{paperpurple}{\textbf{Task-aware adaptation.} Reusing an architectural design across imaging modalities requires aligning feature interfaces, input information and prediction targets. Image--clinical fusion illustrates this adaptation principle in ultrasound.}
Table~\ref{tab:breast} follows Table~\ref{tab:comparison}'s fixed epoch-50 and Tail definitions. Loss is cross entropy averaged over batch means. The 1.96-point endpoint lead over TransXNet represents one additional correct classification among 51 patients (41 versus 40). The Tail lead over RepViT is approximately 7.1 points.

Baseline rankings differed across summaries. TransXNet had higher endpoint accuracy than RepViT, but lower Tail accuracy (73.14\% versus 74.71\%) and higher endpoint loss (0.9303 versus 0.8783). Reporting all three metrics distinguishes endpoint classification, late-training behavior and cross-entropy errors. Tail repeatedly evaluates the same 51 patients across epochs; it does not represent ten independent experiments.

\begin{table}[!t]
\caption{BrEaST multimodal validation. Metrics follow Table~\ref{tab:comparison}; $\dagger$ marks the adapted implementation in Section~\ref{sec:breast_protocol}.}
\label{tab:breast}
\centering\small
\setlength{\tabcolsep}{4pt}
\begin{tabular}{lrrr}
\toprule
Model & Acc. (\%) & Tail (\%) & Val. loss\\
\midrule
EdgeNeXt & 70.59 & 70.59 & 3.1660\\
RepViT & 76.47 & 74.71 & 0.8783\\
TransXNet & 78.43 & 73.14 & 0.9303\\
\textbf{M3D-Net}$^\dagger$ & \textbf{80.39} & \textbf{81.76} & \textbf{0.4297}\\
\bottomrule
\end{tabular}
\end{table}

\textbf{BrEaST scores evaluate complete image--clinical systems.} Image-only and clinical-only controls would separate feature-source contributions. Future experiments should share patient-level splits across models, select configurations using only training and validation subsets, and report every predefined split and seed. This would characterize partition and training variability without selecting runs by held-out accuracy or treating repeated validation measurements as independent evidence.

\section{Discussion}
\textbf{The operators address complementary feature-processing roles.} MUDD retrieves recent features, MCA weights local--global context, and DA contrasts global maps. Fixed-stage MCA removal, static/dynamic MUDD weights and histories, and capacity-matched standard attention would test these roles separately.

DA's zero-sum map concerns only value aggregation; residual paths preserve other information. This property does not establish localization or artifact suppression, and lower loss alone does not establish calibration~\cite{guo2017}. The BrEaST projection shortcut also prevents extending constant-value rejection to the full mixer.

\begin{figure}[!t]
\centering
\includegraphics[width=0.99\linewidth,trim=0bp 8.4bp 0bp 6.5bp,clip]{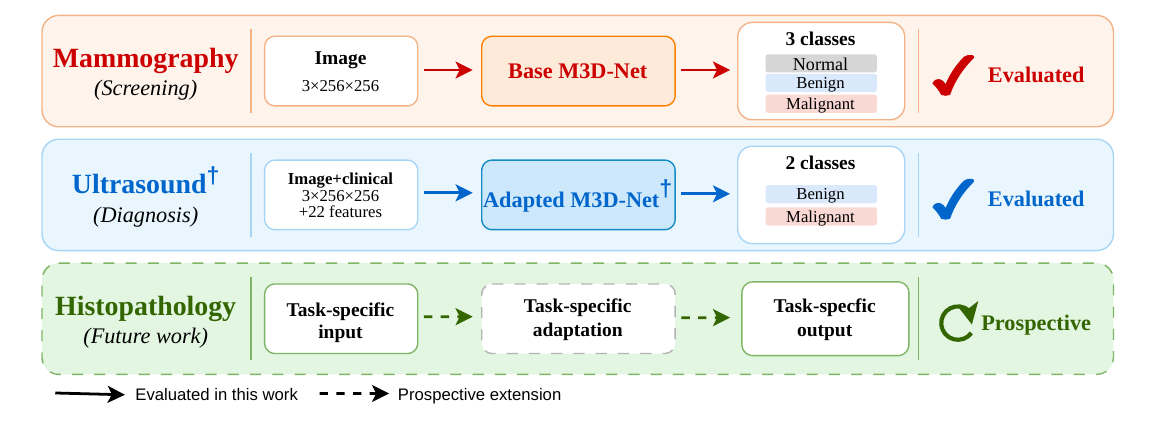}
\caption{Evidence scope of M3D-Net. Solid paths denote evaluated mammography and adapted image--clinical ultrasound settings; the dashed path denotes a prospective, unevaluated histopathology extension. $\dagger$: ultrasound adaptation (Section~\ref{sec:breast_protocol}).}
\label{fig:task_scope}
\end{figure}

Figure~\ref{fig:task_scope} distinguishes evaluated tasks from a potential extension. Ultrasound adaptation changes modality, prediction endpoint and clinical inputs. Transfer benefits remain task-dependent~\cite{raghu2019}; cross-site evaluation~\cite{zech2018} and application to histopathology~\cite{bejnordi2017} require appropriate prediction units and patient-disjoint testing.

\section{Conclusion}
M3D-Net organizes spatial context weighting, bounded feature reuse, and differential aggregation within a \textbf{resolution-aware hierarchy}, linking stage-specific operator placement with structured interactions inside each encoder block. It achieves the highest recorded validation accuracy among the compared mammography backbones, while an adapted ultrasound implementation extends evaluation to image--clinical fusion. The design could extend to related imaging domains such as histopathology through task-specific adaptation. Repeated-seed, component-controlled and independent evaluations are needed to characterize performance beyond the reported protocols.

\pagebreak[4]
\clearpage
\balance
\section{Compliance with Ethical Standards}
We reuse public, de-identified AISSLab and BrEaST data (CC BY 4.0)~\cite{kaggle_aisslab,pawlowska2024breast}. AISSLab's source reports approval SUIRB-HR-2023-013 from the Al-Ma'amon Diagnostic Centre IRB~\cite{aisslab2026}. BrEaST's original collection was approved by the Lower Silesian Chamber of Medicine Bioethics Committee (2/BNR/2022), with written consent waived for retrospective anonymized data~\cite{pawlowska2024breast}. No new participants were recruited or primary data collected.
\section{Funding Acknowledgement}
This work was supported by the Longgang District Special Funds for Science and Technology Innovation under Grant \mbox{LGKCSDPT20250}.
The authors declare no conflicts of interest.


\begin{thebibliography}{99}
\setlength{\itemsep}{3.85pt plus .4pt minus .2pt}
\bibitem{geras2017} K. J. Geras \emph{et al.}, ``High-resolution breast cancer screening with multi-view deep convolutional neural networks,'' 2017, \href{https://arxiv.org/abs/1703.07047}{arXiv:1703.07047}.
\bibitem{mckinney2020} S. M. McKinney \emph{et al.}, ``International evaluation of an AI system for breast cancer screening,'' \emph{Nature}, vol. 577, pp. 89--94, 2020, doi: \doi{10.1038/s41586-019-1799-6}.
\bibitem{lotter2021} W. Lotter \emph{et al.}, ``Robust breast cancer detection in mammography and digital breast tomosynthesis using an annotation-efficient deep learning approach,'' \emph{Nat. Med.}, vol. 27, pp. 244--249, 2021, doi: \doi{10.1038/s41591-020-01174-9}.
\bibitem{wu2020} N. Wu \emph{et al.}, ``Deep neural networks improve radiologists' performance in breast cancer screening,'' \emph{IEEE Trans. Med. Imag.}, vol. 39, no. 4, pp. 1184--1194, 2020, doi: \doi{10.1109/TMI.2019.2945514}.
\bibitem{shen2019} L. Shen, L. R. Margolies, J. H. Rothstein, E. Fluder, R. McBride, and W. Sieh, ``Deep learning to improve breast cancer detection on screening mammography,'' \emph{Sci. Rep.}, vol. 9, Art. no. 12495, 2019, doi: \doi{10.1038/s41598-019-48995-4}.
\bibitem{eisemann2025} N. Eisemann \emph{et al.}, ``Nationwide real-world implementation of AI for cancer detection in population-based mammography screening,'' \emph{Nat. Med.}, vol. 31, pp. 917--924, 2025, doi: \doi{10.1038/s41591-024-03408-6}.
\bibitem{lou2025} M. Lou, S. Zhang, H.-Y. Zhou, S. Yang, C. Wu, and Y. Yu, ``TransXNet: Learning both global and local dynamics with a dual dynamic token mixer for visual recognition,'' \emph{IEEE Trans. Neural Netw. Learn. Syst.}, vol. 36, no. 6, pp.~\mbox{11534--11547, 2025}, doi: \doi{10.1109/TNNLS.2025.3550979}.
\bibitem{maaz2022} M. Maaz, A. Shaker, H. Cholakkal, S. Khan, S. W. Zamir, R. M. Anwer, and F. S. Khan, ``EdgeNeXt: Efficiently amalgamated CNN--Transformer architecture for mobile vision applications,'' in \emph{Proc. ECCV 2022 Workshops}, pp. 3--20, 2023, doi: \doi{10.1007/978-3-031-25082-8_1}.
\bibitem{wang2024} A. Wang, H. Chen, Z. Lin, J. Han, and G. Ding, ``RepViT: Revisiting mobile CNN from ViT perspective,'' in \emph{Proc. IEEE/CVF CVPR}, pp. 15909--15920, 2024, doi: \doi{10.1109/CVPR52733.2024.01506}.
\bibitem{sanghvi2026} S. Sanghvi, P. Miglani, S. Shashikumar, K. R. Borgavi, V. Singla, and C. Arora, ``Attend what matters: Leveraging vision foundational models for breast cancer classification using mammograms,'' 2026, \href{https://arxiv.org/abs/2604.19350}{arXiv:2604.19350}.
\bibitem{xiao2025} D. Xiao, Q. Meng, S. Li, and X. Yuan, ``MUDDFormer: Breaking residual bottlenecks in Transformers via multiway dynamic dense connections,'' in \emph{Proc. ICML}, PMLR 267, pp.~\mbox{68440--68458, 2025}. [Online]. Available: \url{https://proceedings.mlr.press/v267/xiao25d.html}.
\bibitem{ye2025} T. Ye, L. Dong, Y. Xia, Y. Sun, Y. Zhu, G. Huang, and F. Wei, ``Differential Transformer,'' in \emph{Proc. ICLR}, 2025. [Online]. Available: \url{https://openreview.net/forum?id=OvoCm1gGhN}.
\bibitem{hou2021} Q. Hou, D. Zhou, and J. Feng, ``Coordinate attention for efficient mobile network design,'' in \emph{Proc. IEEE/CVF CVPR}, pp. 13708--13717, 2021, doi: \doi{10.1109/CVPR46437.2021.01350}.
\bibitem{huang2017} G. Huang, Z. Liu, L. van der Maaten, and K. Q. Weinberger, ``Densely connected convolutional networks,'' in \emph{Proc. IEEE CVPR}, pp. 2261--2269, 2017, doi: \doi{10.1109/CVPR.2017.243}.
\bibitem{he2016} K. He, X. Zhang, S. Ren, and J. Sun, ``Deep residual learning for image recognition,'' in \emph{Proc. IEEE CVPR}, pp.~\mbox{770--778, 2016}, doi: \doi{10.1109/CVPR.2016.90}.
\bibitem{zhao2017} H. Zhao, J. Shi, X. Qi, X. Wang, and J. Jia, ``Pyramid scene parsing network,'' in \emph{Proc. IEEE CVPR}, pp. 6230--6239, 2017, doi: \doi{10.1109/CVPR.2017.660}.
\bibitem{kaggle_aisslab} ``AISSLab Breast Cancer Dataset,'' Kaggle. Accessed: Sep. 5, 2026. [Online]. Available: \url{https://www.kaggle.com/datasets/orvile/aisslab-breast-cancer-dataset}.
\bibitem{aisslab2026} A. M. Al-Hejri, M. Fazea, A. H. Sable, R. M. Al-Tam, M. A. Al-antari, and S. S. Alshamrani, ``Toward general AI harmonization with real mammogram imaging breast cancer dataset,'' \emph{BMC Res. Notes}, vol. 19, Art. no. 190, 2026, doi: \doi{10.1186/s13104-026-07727-4}.
\bibitem{pawlowska2024breast} A. Paw{\l}owska \emph{et al.}, ``Curated benchmark dataset for ultrasound based breast lesion analysis,'' \emph{Sci. Data}, vol. 11, Art. no. 148, 2024, doi: \doi{10.1038/s41597-024-02984-z}.
\bibitem{liu2024itransformer} Y. Liu, T. Hu, H. Zhang, H. Wu, S. Wang, L. Ma, and M. Long, ``iTransformer: Inverted Transformers are effective for time series forecasting,'' in \emph{Proc. ICLR}, 2024. [Online]. Available: \url{https://openreview.net/forum?id=JePfAI8fah}.
\bibitem{guo2017} C. Guo, G. Pleiss, Y. Sun, and K. Q. Weinberger, ``On calibration of modern neural networks,'' in \emph{Proc. ICML}, PMLR 70, pp. 1321--1330, 2017. [Online]. Available: \url{https://proceedings.mlr.press/v70/guo17a.html}.
\bibitem{raghu2019} M. Raghu, C. Zhang, J. Kleinberg, and S. Bengio, ``Transfusion: Understanding transfer learning for medical imaging,'' in \emph{Adv. Neural Inf. Process. Syst.}, vol. 32, 2019. [Online]. Available: \url{https://papers.nips.cc/paper/2019/hash/eb1e78328c46506b46a4ac4a1e378b91-Abstract.html}.
\bibitem{zech2018} J. R. Zech, M. A. Badgeley, M. Liu, A. B. Costa, J. J. Titano, and E. K. Oermann, ``Variable generalization performance of a deep learning model to detect pneumonia in chest radiographs: A cross-sectional study,'' \emph{PLoS Med.}, vol. 15, no. 11, Art. no. e1002683, 2018, doi: \doi{10.1371/journal.pmed.1002683}.
\bibitem{bejnordi2017} B. Ehteshami Bejnordi \emph{et al.}, ``Diagnostic assessment of deep learning algorithms for detection of lymph node metastases in women with breast cancer,'' \emph{JAMA}, vol. 318, no. 22, pp. 2199--2210, 2017, doi: \doi{10.1001/jama.2017.14585}.
\end{thebibliography}
\end{document}